\documentclass[runningheads]{llncs}
\usepackage[T1]{fontenc}
\usepackage{graphicx}
\usepackage{graphicx}
\usepackage{xcolor}
\usepackage{comment}
\usepackage[utf8]{inputenc}
\usepackage{multirow}
\usepackage{arydshln}
\usepackage{subfigure}
\usepackage{float}
\usepackage{array}
\usepackage{tabularx}
\usepackage{graphicx}
\usepackage{url}
\usepackage{adjustbox}
\usepackage{xspace}
\usepackage{xcolor}
\usepackage{comment}
\usepackage{caption} 
\usepackage{booktabs}
\usepackage{hyperref}
\PassOptionsToPackage{bookmarks=false}{hyperref}
\usepackage{nicematrix}
\restylefloat{table}
\usepackage{latexsym}
\usepackage{amsmath, amssymb}
\usepackage{type1cm}        
\usepackage{comment}
\usepackage{makeidx}         
\usepackage{multicol}        
\usepackage[bottom]{footmisc}
\usepackage{newtxtext}       %
\usepackage[varvw]{newtxmath} 
\usepackage{type1cm}        
\usepackage{makeidx}         
\usepackage{graphicx}        
\usepackage{multicol}        
\usepackage[bottom]{footmisc}
\usepackage{newtxtext}       %
\usepackage[varvw]{newtxmath}       

\definecolor{backgG}{RGB}{255, 255, 153}
\definecolor{tagtxtG}{RGB}{102, 102, 0}
\definecolor{backgPc}{RGB}{179, 255, 179}
\definecolor{tagtxtPc}{RGB}{0, 102, 0}
\definecolor{backgPw}{RGB}{255, 179, 179}
\definecolor{backgPw}{rgb}{0.0, 1.0, 1.0}
\definecolor{tagtxtPw}{RGB}{0.0, 1.0, 1.0}
\definecolor{backgPo}{rgb}{0.76, 1, 1}
\definecolor{tagtxtPo}{RGB}{102, 0, 0}
\definecolor{backgPm}{rgb}{0.98, 0.81, 0.69}
\definecolor{tagtxtPm}{RGB}{0,1,1}

\makeindex 

\begin{document}
\title{Analysis of Multidomain Abstractive Summarization Using Salience Allocation}
%
%
\author{Tohida Rehman\inst{1}\thanks{corresponding author} \and
Raghubir Bose \inst{2}\and Soumik Dey\inst{1}\and\\ Samiran Chattopadhyay\inst{3,1}}

\authorrunning{Rehman, et al.}
\titlerunning{Rehman, et al.}

\institute{Jadavpur University, Kolkata, India.\\ \email{\{tohida.rehman,soumikdey525\}@gmail.com}
\and  BFSI BTAG , Tata Consultancy Services.\\
\email{raghubir.bose@tcs.com} \and Techno India University, Kolkata, India\\
\email{samirancju@gmail.com}}

\maketitle
\begin{abstract}\unskip
This paper explores the realm of abstractive text summarization through the lens of the SEASON (Salience Allocation as Guidance for Abstractive SummarizatiON) technique, a model designed to enhance summarization by leveraging salience allocation techniques. The study evaluates SEASON's efficacy by comparing it with prominent models like BART, PEGASUS, and ProphetNet, all fine-tuned for various text summarization tasks. The assessment is conducted using diverse datasets including CNN/Dailymail, SAMSum, and Financial-news based Event-Driven Trading (EDT), with a specific focus on a financial dataset containing a substantial volume of news articles from 2020/03/01 to 2021/05/06.
This paper employs various evaluation metrics such as ROUGE, METEOR, BERTScore, and MoverScore to evaluate the performance of these models fine-tuned  for generating abstractive summaries. 
The analysis of these metrics offers a thorough insight into the strengths and weaknesses demonstrated by each model in summarizing news dataset, dialogue dataset and financial text dataset. 
The results presented in this paper not only contribute to the evaluation of the SEASON model's effectiveness but also illuminate the intricacies of salience allocation techniques across various types of datasets.


\end{abstract}
\section{Introduction}
Given the abundance of information in various domains, it becomes essential for humans to focus on remembering only the key summary points.
There is a demand for a solution capable of delivering accurate and timely summary data, requiring a tool or approach to extract precise summaries from extensive datasets. Automatic text summarization serves as an instance of a technology that can be employed for this purpose. Generally, summarization can be categorized into two approaches: extractive and abstractive \cite{luhn1958automatic}.
Extractive summarization entails the process of choosing and merging sentences or passages from the source text to generate a concise version that preserves the essential information.
Abstractive summarization, models traditionally undergo end-to-end training using extensive datasets comprising raw text alongside human-generated summaries. Recent research efforts have focused on integrating extractive salience guidance into abstractive summarization models to enhance their comprehension of input documents. Relying exclusively on extractive summaries as guiding references can lead to the omission of vital information or a lack of coherence in the ultimate summary. 
The primary contributions outlined in this paper are as follows::
\begin{enumerate}
     \item This paper incorporates an abstractive summarization technique, SEASON(SaliencE Allocation as Guidance for Abstractive SummarizatiON), proposed by Wang et al. \cite{sd1}.
     The SEASON model's provides an effective framework for accentuating crucial elements of text summarization. Salience, indicating a sentence's contribution to the primary context of a document, is evaluated for its allocation across the document's sentences. This evaluation involves a linear classifier overlaying the encoder, integrating Salience-Aware Cross-Attention (SACA) within the decoder to guide the abstractive summarization process.
   \item This study evaluates the SEASON model's performance using established Financial-news based Event-Driven Trading (EDT) datasets\cite{zhou-etal-2021-trade} represents financial text, others notably CNN/Dailymail \cite{sd3a} and SAMSum \cite{sd4} datasets.   
   \item Furthermore, it compares SEASON's performance with three pre-trained models namely, BART, PEGASUS, and ProphetNet—specifically fine-tuned for three datasets. The experimental results illustrate the competitive edge of the SEASON model, surpassing BART, PEGASUS, and ProphetNet on three datasets across various evaluation metrics such as ROUGE \cite{sd27}, METEOR \cite{sd28}, MoverScore \cite{sd29}, and BERTScore \cite{sd30}.  

\end{enumerate}
\section{Literature Review} 
Nallapati et al.\cite{sd3a} introduced a sequence-to-sequence framework for generating abstractive summaries using bidirectional and unidirectional LSTMs, plus attention. See et al. \cite{See2017GetTT} introduced a Pointer-Generator Network with coverage, aiming to enhance accuracy and reduce redundancy. 
The pointer-generator model, coupled with a coverage mechanism, addresses the issue of out-of-vocabulary (OOV) words and mitigates the problem of generating repetitive phrases, as integrating in \cite{rehman2021automatic,rehman-etal-2022-named,rehman2023research,10172215}. 
Vaswani et al. \cite{sd13} revolutionized NLP with the transformer model, using self-attention to capture long-range dependencies and enable efficient parallel training.
The implementation of the transformer architecture, particularly the bidirectional encoder model BERT, has demonstrated improved performance in downstream NLP tasks, including text summarization \cite{devlin2018bert,rehman2022transfer}.
PEGASUS is a pre-trained model designed for summarization tasks, trained on large corpora and incorporating a gap sentence generation task across 12 downstream summarization tasks \cite{zhang2020pegasus}.
Aksenov et al. \cite{aksenov2020abstractive} proposed the BERT windowing method as a solution to overcome the limitations of input size in BERT-based architectures.
Hsu et al. \cite{sd14} proposed an integrated summarization model combining extractive and abstractive techniques, achieving high ROUGE scores and readability.
Gehrmann et al. \cite{sd15} suggested a bottom-up abstractive summarization method with a data-efficient content selector that improves summarization performance on CNN/Dailymail and  NYT datasets, outperforming other content selection models.
Li et al. \cite{sd16} proposed a guiding generation model that merges extractive and abstractive methods, using keywords and a prediction-guide mechanism, improving summary production significantly on the CNN/Dailymail dataset.
Jin et al. \cite{sd17} introduced SemSUM, a summarization model integrating semantic dependency graphs to enhance abstractive summarization, using a sentence encoder, graph encoder, and summary decoder.
Zhu et al. \cite{sd18} enhanced abstractive summarization accuracy by integrating knowledge graphs, improving factual correctness and ROUGE scores on the CNN/Dailymail dataset.
Saito et al. \cite{sd19} suggested a summarization model combining a saliency model and a sequence-to-sequence model to enhance abstractive summarization performance.
Manakul and Gales \cite{sd20} proposed an innovative summarization approach, merging local self-attention and extractive summarization, improving accuracy in summarizing lengthy documents.
Recent research in summarization highlights that integrating extractive summaries into input documents boosts abstractive summarization efficiency, improving accuracy and coherence \cite{sd19,sd21}.
Rehman at al. \cite{rehman2022analysis} analyzed various pre-trained models, including google/pegasus-cnn-dailymail, T5-base, and facebook/bart-large-cnn, for text summarization across diverse datasets such as CNN-dailymail, SAMSum, and BillSum. 
Cao et al. \cite{sd22} introduced a novel summarization method, selecting relevant training data to enhance summarization quality and coherence using appropriate templates to reduce hallucination.
Recent studies shown, \cite{nan-etal-2021-entity,rehman2023hallucination} hallucination can be reduced by employing a filter on the training data and utilizing multi-task learning techniques. 

Recent studies, investigate using selective attention in NLP models to incorporate prior information, improving summarization, especially at the sentence level \cite{sd25,wang2022robust}.
Wang et al. \cite{sd1} introduced SEASON, a model using salience-guided selective attention to enhance abstractive summarization performance effectively.

\section{Pre-trained Models}

We have used four distinct transformer-based pre-trained models, SEASON, BART, PEGASUS, and ProphetNet, demonstrate unique architectural paradigms for abstractive summarization.

\textbf{SEASON} (Salience Allocation as Guidance for Abstractive SummarizatiON) model is built upon a conventional transformer-based model designed for abstractive summarization \cite{sd1}. It integrates salience prediction and text summarization within a unified network. It encapsulates salience prediction and text summarization in a single network. 
This integration involves a salience predictor within the encoder, associating expected salience levels with corresponding embeddings. These salience embeddings are then fused with the key vectors in the cross-attention mechanism. Multi-task end-to-end training is conducted, and inference is achieved in a single forward pass. Throughout the training process, the model is taught to estimate the level of salience assigned to each sentence,  by salience allocation determined through ROUGE-based benchmarks, the model generates the abstractive summary. In the inference phase, for generating the summary, the SEASON model predicts expected salience allocation using encoder outputs and uses this information to guide the decoder in generating the summary.

\textbf{BART} is referred to as the Bidirectional and Auto-Regressive Transformers model  \cite{sd5}.
This model encompasses a Transformer encoder-decoder structure that merges a bidirectional encoder, reminiscent of BERT with an autoregressive decoder.
The input sequence is initially processed with a bidirectional encoder trained on intentionally corrupted text.
The encoder randomly masks certain words in the input, and then the left-to-right autoregressive decoder generates the output, relying on the encoded input.
After extensive training on a large corpus, BART becomes proficient at predicting missing tokens.
This proficiency allows it to excel at restoring corrupted real-time sentences, and it can be further fine-tuned for various NLP tasks.

In natural language processing (NLP), a pre-trained model \textbf{PEGASUS} offers a novel method for abstractive summarization \cite{sd6}. 
It uses a contextual token prediction approach for self-supervised pre-training to understand language structure and semantics without labeled data.
PEGASUS, akin to BERT, employs a revised MLM approach for self-supervised pre-training. PEGASUS introduces Gap Sentences as a training objective.
Random input sentences are replaced with \textit{[MASK]} tokens, and then the model predicts them based on the remaining context.
After pre-training, PEGASUS is fine-tuned for abstractive summarization using supervised learning. It achieves top-tier performance on 12 benchmark datasets.

\textbf{ProphetNet}, a transformer-based NLP model, aims to overcome limitations in existing transformer models \cite{sd7}.
It introduces future n-gram prediction, forecasting upcoming tokens using context, and improving its ability to handle long sequences.
It employs n-stream self-attention mechanism, enhancing relevance-focused attention and overall performance.
It adopts a mask-centered autoencoder denoising task, training the model to restore obscured input tokens, improving robustness against noisy data. The alterations introduced by ProphetNet to the Transformer architecture position it as a potent and efficient model capable of addressing a broad spectrum of Natural Language Processing (NLP) tasks.

\begin{figure}[t!]
  \centering
  \includegraphics[width=.80\textwidth]{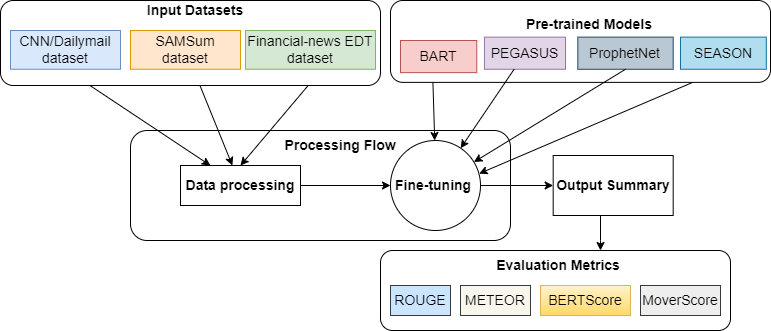}
  \caption{Workflow of our system}
  \label{fig:workflow}
\end{figure}

Figure \ref{fig:workflow} illustrates system's architecture using a detailed flow diagram, providing a clear understanding of its smooth and seamless functionality.

\section{Experimental Setup}
\subsection{Datasets}
This study employs the \textbf{CNN/Dailymail} \cite{sd3a}, \textbf{SAMSum} \cite{sd4}, and \textbf{Financial-news based Event-Driven Trading (EDT)} \cite{zhou-etal-2021-trade} datasets for text summarization tasks. The CNN/ Dailymail dataset version 3.0.0 contains over 3,00,000 news articles, each paired with handcrafted summaries and highlights from Daily Mail and CNN writers. SAMSum comprises 16,000 chat dialogues with human-generated summaries. The Financial-news based Event-Driven Trading (EDT) dataset includes 921 news articles with event and token labels and over 3,00,000 articles with comprehensive stock price labels, serving as a key corpus for financial domain summarization. It spans from March 1, 2020, to May 6, 2021, drawn from reputable platforms like PRNewswire and Businesswire. Due to limitations in computational resources, a subset of 5,000 training, 625 validation, and 625 testing data points was employed from each datasets. 
This selection facilitates experimentation within computational constraints, offering a diverse range of data sources for effective model training and evaluation.

\subsection{Implementation Details}
In the process of preparing the training data, we introduce a unique token before starting every sentence and computing its representation. Additionally, we limit the length of each input sequence in the CNN/Dailymail, SAMSum, and Financial-news based Event-Driven Trading (EDT) datasets to 512, 256, and 512 tokens, respectively, including special tokens. To maintain the essence of the reference summaries for each datasets, we limit the length of each target sequence to 100, 50, and 40 tokens, respectively. 

For fine-tuning the BART model, we take the pre-trained \textit{facebook/bart-large-cnn} model from the hugging face website \footnote{\url{https://huggingface.co/facebook/bart-large-cnn}}. For fine-tuning the PEGASUS model, we take a pre-trained \textit{google/pegasus-cnn\_dailymail} model from the hugging face website \footnote{\url{https://huggingface.co/google/pegasus-cnn_dailymail}}. For fine-tuning the ProphetNet model, we take a pre-trained \textit{microsoft/prophetnet-large-uncased} model from the hugging face website\footnote{\url{https://huggingface.co/microsoft/prophetnet-large-uncasedl}}.
For the SEASON model, we fine-tune the pre-trained \textit{facebook/bart-large} model from the hugging face website \footnote{\url{https://huggingface.co/facebook/bart-large}}. 

For measuring salience, we utilize ROUGE-L F1 metric. During inference, we utilize predicted soft estimation to allocate expected salience, employing a temperature of 0.5 to sharpen the probability of salience degree. Our approach for inference involves using beam search, where the beam width is set to 5, applying a length penalty set to 1.5, and implementing 3-gram blocking. We have conducted fine-tuning for each of the models throughout 5 epochs only. We trained all models on Tesla T4 \texttt{Colab } that supports GPU-based training. 

\subsection{Evaluation Methods}
Following the summarization process by fine-tuned of pre-trained models, the generated summaries undergo evaluation using ROUGE, METEOR, BERTScore, and MoverScore metrics.
ROUGE \cite{sd27} metric evaluate the matched model predicted summary to human written summary, measuring n-gram overlap (ROUGE-N) and longest common sequence (ROUGE-L). ROUGE-Lsum offers detailed insights for enhancing NLP summarization task. 
METEOR  metric \cite{sd28} evaluates the match between model predicted summary to human written summar by assigning a score that considers a blend of unigram precision, unigram recall, and a fragmentation measure. This measure is designed to accurately reflect the degree of ordering alignment between the matched words.
MoverScore \cite{sd29} metric emphasizes faithfulness by measuring text deviations from references, while BERTScore \cite{sd30} uses BERT embeddings for semantic and contextual assessment, overcoming traditional metric limitations.
\section{Results}
\subsection{Comparison of Fine-tuned Models Performance}
In this work, we have used ROUGE-N (ROUGE-1, ROUGE-2), ROUGE-L, ROUGE-Lsum, METEOR, BERTScore, and MoverScore metrics to compare the model performance for the CNN/Dailymail, SAMSum, and Financial-news based Event-Driven Trading (EDT) datasets selected for test datasets. 

The Table \ref{Table:par_all_types_rouge_meteor_bert} shows the evaluation metric results of the four models on the three datasets. The fine-tuned SEASON model gives better scores in all metrics using Financial-news based Event-Driven Trading (EDT) datasets. The SEASON model fine-tuned on the SAMSum dataset gives better scores in terms of ROUGE-1, ROUGE-2, ROUGE-L, METEOR, BERTScore, and MoverScore while the fine-tuned ProphetNet model receives higher scores in respect of the ROUGE-Lsum metric only.
The PEGASUS model fine-tuned on CNN/Dailymail gives higher scores in respect of all metrics.
Results show SEASON model outperforms the other models across multiple evaluation metrics  on major datasets, indicating its competitive advantage towards abstractive summarization.

\begin{table*}[!htbp]
    \centering
    \caption{Comparison of the SEASON models with others three fine-tuned models BART, PEGASUS, and ProphetNet: F1-scores for ROUGE, METEOR, BERTScore, and MoverScore on different datasets. All scores in percentage (\%).}
    \label{Table:par_all_types_rouge_meteor_bert}
    \begin{adjustbox}{width=.95\linewidth}
        \begin{tabular}{|p{2.4cm}p{2cm}ccccccc|}  
            \hline
            & & ROUGE-1 & ROUGE-2 & ROUGE-L & ROUGE-Lsum & METEOR & BERTScore & MoverScore \\
            Dataset & Model Name & F1 & F1 & F1 & F1 & Final score & F1 & F1 \\ \hline
            \multirow{4}{4em}{CNN/Dailymail} & BART & 34.85 & 13.78 & 24.26 & 31.93 & 24.43 & 88.27 & 12.50 \\
            & PEGASUS & \bf{35.89} & \bf{15.52} & \bf{26.39} & \bf{32.37} & \bf{29.34} & \bf{88.58} & \bf{13.67} \\
            & ProphetNet & 33.82 & 13.22 & 23.75 & 30.78 & 24.44 & 87.23 & 11.68 \\
            & SEASON & 34.65 & 13.89 & 24.25 & 24.25 & 24.77 & 87.94 & 12.41 \\ \hline
            \multirow{4}{4em}{SAMSum} & BART & 45.62 & 21.53 & 35.15 & 41.51 & 33.40 & 90.58 & 25.76 \\
            & PEGASUS & 45.13 & 21.10 & 35.06 & 39.47 & 33.90 & 90.52 & 26.02 \\
            & ProphetNet & 50.02 & 25.08 & 39.68 & \bf{43.86} & 40.26 & 89.49 & 30.66 \\
            & SEASON & \bf{50.61} & \bf{25.72} & \bf{42.25} & 42.35 & \bf{49.63} & \bf{91.82} & \bf{31.96} \\ \hline
            \multirow{4}{4em}{Financial-news based EDT} & BART & 47.15 & 28.27 & 41.10 & 41.14 & 35.76 & 88.91 & 28.15 \\
            & PEGASUS & 43.16 & 25.13 & 37.14 & 37.15 & 31.61 & 88.13 & 22.45 \\
            & ProphetNet & 50.63 & 31.21 & 44.67 & 44.77 & 38.97 & 87.18 & 32.02 \\
            & SEASON & \bf{52.91} & \bf{34.64} & \bf{48.15} & \bf{48.15} & \bf{51.20} & \bf{90.58} & \bf{35.43} \\ \hline
        \end{tabular}
    \end{adjustbox}
\end{table*}

\subsection{Sample Output}

Tables \ref{output_cnndm}, \ref{output_samsum}, and \ref{output_financial} illustrated the outputs generated by fine-tuned four models on different datasets.
We have used the \colorbox{backgG}{yellow} color to highlight the hallucinated words/phrases generated by any models.
We observed in table \ref{output_cnndm}, that the fine-tuned SEASON model generates a better meaningful summary than other fine-tuned models.
In table \ref{output_samsum}, we observed that fine-tuned the SEASON model generates a better summary than PEGASUS, ProphetNet, and BART on the SAMSUM dataset. Summaries produced by PEGASUS and ProphetNet fine-tuned models lack conciseness and consist of hallucinated facts like \colorbox{backgG}{70 tickets} and \colorbox{backgG}{7 pm}.
In table \ref{output_financial}, we observed that the fine-tuned SEASON model generates a better summary than the other three models on the  Event-Driven Trading (EDT) dataset.
\begin{table}[h!] 
\centering 
\begin{tabular}{ |p{12.2cm}|} \hline
\textbf{Input:} ``(CNN) Singer-songwriter Joni Mitchell has been hospitalized in 
$\cdot ..................... $ 
71, was inducted into the Rock and Roll Hall of Fame in 1997." \\ \hline

\textbf{Ground-truth Summary:}  ``Mitchell is best known for her hits "Big Yellow Taxi" and "Free Man in Paris". Paramedics came to her Los Angeles home on Tuesday afternoon." \\ \hline
\textbf {Fine-tuned BART:} ``Joni Mitchell is hospitalized in Los Angeles. "She regained consciousness on the ambulance ride to an L.A. area hospital," her website says. Some of her best-known songs are "Big Yellow Taxi," "Help Me" and "Free Man in Paris". She was inducted into the Rock and Roll Hall of Fame in 1997." \\ \hline
\textbf {Fine-tuned PEGASUS:}  ``Singer-songwriter Joni Mitchell has been hospitalized and is in intensive care. Mitchell, 71, was inducted into the Rock and Roll Hall of Fame in 1997."
 \\ \hline
\textbf {Fine-tuned ProphetNet:} ``Joni Mitchell was found unconscious in her home. She regained consciousness on the ambulance ride to the hospital, spokesman said. The 71-year-old has written hits for other artists too. Mitchell was inducted into the Rock and Roll Hall of Fame in 1997."  \\ 
 \hline
 \textbf {Predicted Summary with SEASON :} ``Joni Mitchell was found unconscious in her Los Angeles home on Tuesday afternoon. She regained consciousness and is in intensive care. Some of her best-known songs are "Big Yellow Taxi," "Help Me" and "Free Man in Paris". "\\ 
 \hline
\end{tabular}
\caption{Comparison of ground-truth and generated summary derived from the test set of the CNN/Dailymail dataset.. For all models, input is taken from the CNN/Dailymail news article: \texttt{\url{https://edition.cnn.com/2015/04/01/entertainment/joni-mitchell-hospitalized/index.html}}}
\label{output_cnndm}
\end{table}

\begin{table}[h!] 
\centering 
\begin{tabular}{ |p{12.2cm}|} \hline
 \textbf{Input:}``Sally: Hey! Imagine Dragons are coming to us!
Tim: So I've heard.\\
Sally: And you didn't tell me?! 
Tim: Come on. It's just a band...\\
Sally: It's not JUST a band, you jerk!
Sally: $<$file\_video$>$ 
Sally: I've already checked the ticket availability. There are still some tickets for the standing area at our ABC Theatre. Shall we go together?\\
Tim: How much are they?
Sally: 70
Tim: When is the gig?
Sally: 12 July
Tim: Well, I may go.
Sally: Your enthusiasm is infectious, really... Try inviting me for some sports events and you'll see how happy I'll be.\\
Tim: Ok! Let's go! It'll be an unforgettable evening!
Sally: Jerk! I've already bought the tickets, so put it in your diary. Tim: Done." 
\\
 \hline
 \textbf{Ground-truth Summary:} ``Imagine Dragons have a concert at ABC Theatre on 12 July. Sally wants to go with Tim. She bought tickets, they cost 70." \\ 
 \hline
 \textbf {Fine-tuned  BART:} ``Sally and Tim will go to the Imagine Dragons concert on 12 July together." \\ 
 \hline
 \textbf {Fine-tuned  PEGASUS:} ``Imagine Dragons are coming to Sally's ABC Theatre on 12 July. There are \colorbox{backgG}{70 tickets} for the standing area at the theatre. Sally has bought the tickets." \\ 
 \hline
 \textbf {Fine-tuned ProphetNet:}  ``Imagine Dragons are coming to the ABC Theatre on 12 July. Sally has bought the tickets. the cost is 70. Tim will go with her. The concert is at \colorbox{backgG}{7 pm}."  \\ 
 \hline
 \textbf {Fine-tuned SEASON :} ``Imagine Dragons are coming to the ABC Theatre on 12 July. Tim and Sally will go together."   \\ 
 \hline
\end{tabular}
\caption{Comparison of ground-truth and generated summary derived from the test set of the SAMSum dataset.}
\label{output_samsum}
\end{table}

\begin{table}[h!] 
\centering 
\begin{tabular}{ |p{12.2cm}|} \hline
 \textbf{Input:} ``NEW BRUNSWICK, N.J.--(BUSINESS WIRE)--Brunswick Bancorp (Brunswick or the Company) (OTC: BRBW), the holding company for Brunswick 
 $\cdot ............................... $ 
 through its New Brunswick main office and four additional branch offices."
\\ \hline
 \textbf{Ground-truth Summary:}  ``Brunswick Bancorp Announces Final Results Of 2021 Annual Meeting"
 \\ 
 \hline
 \textbf {Fine-tuned BART:} ``Brunswick Bancorp Announces the Results of the Vote Count for the 2021 Annual Meeting of Shareholders and the Election of James Atieh and Nicholas Frungillo"
 \\ 
 \hline
 \textbf {Fine-tuned PEGASUS:} ``Brunswick Bancorp Announces Re-election of James Atieh and Nicholas A. Frungillo, Jr. to the Board of Directors at 2021 Annual Meeting of Shareholders"
 \\ 
 \hline
 \textbf {Fine-tuned ProphetNet:} ``Brunswick Bancorp announces the re-election of James Atieh and Nicholas A. Frungillo to the Board of Directors"
 \\ 
 \hline
 \textbf {Fine-tuned SEASON :} ``Brunswick Bancorp Announces Results of 2021 Annual Meeting of Shareholders"
  \\ 
 \hline
\end{tabular}
\caption{Comparison of ground-truth and generated summary derived from the test set of the Financial-news based Event-Driven Trading (EDT) dataset.  For all models, input is taken from the EDT Financial-news article: \texttt{\url{https://www.bloomberg.com/press-releases/2021-04-29/brunswick-bancorp-announces-final-results-of-2021-annual-meeting}}}
\label{output_financial}
\end{table}

\section{Conclusion and Future Work}
The paper extensively evaluates abstractive summarization methods using salience allocation, focusing on SEASON, BART, PEGASUS, and ProphetNet fine-tuned models across three datasets. SEASON's fine-tuned model salience allocation significantly improves the summarization by emphasizing crucial details. 
In the future, the salience allocation method's performance should be tested on a larger dataset to assess scalability and generalizability. 
Additionally, the method can be extended to summarize multiple scholarly documents, comparing its effectiveness with other advanced summarization techniques. Furthermore, it has potential applications in various text generation tasks, including question-answering, and chatbot development.

\bibliographystyle{unsrt}
\bibliography{sd,anthology}
\end{document}